# A Study on Quantifying Sim2Real Image Gap in Autonomous Driving Simulations Using Lane Segmentation Attention Map Similarity

**Seongjeong Park[1], Jinu Pahk[1], Lennart Lorenz Freimuth Jahn [1], Yongseob Lim[1], Jinung An[1], Gyeungho Choi[1]**

[1] Daegu Gyeongbuk Institute of Science and Technology, Daegu, Korea
{4or2psj, kyg132, lennart, yslim73, robot, ghchoi}@dgist.ac.kr

* Seongjeong Park and Jinu Pahk contributed equally to this work

**Abstract.** Autonomous driving simulations require highly realistic images. Our preliminary study found that when the CARLA Simulator image was made more like reality by using DCLGAN, the performance of the lane recognition model improved to levels comparable to real-world driving. It was also confirmed that the vehicle's ability to return to the center of the lane after deviating from it improved significantly. However, there is currently no agreed-upon metric for quantitatively evaluating the realism of simulation images. To address this issue, based on the idea that FID (Fréchet Inception Distance) measures the feature vector distribution distance using a pre-trained model, this paper proposes a metric that measures the similarity of simulation road images using the attention map from the self-attention distillation process of ENet-SAD. Finally, this paper verified the suitability of the measurement method by applying it to the image of the CARLA map that implemented a real-world autonomous driving test road.

## 1 Introduction

Simulation is essential to test autonomous driving algorithms in extreme weather conditions and dangerous driving scenarios. [1] Specifically, in the context of autonomous driving, adverse weather conditions can lead to hazardous scenarios, such as a decline in lane recognition accuracy. Therefore, it is crucial to develop a realistic simulation environment that effectively captures and models these challenging situations for the enhancement of autonomous vehicle performance. However, the quantitative evaluation index has not yet been formally agreed upon. In our previous paper, we showed that the Fréchet Inception Distance (FID) score measures the relative realism of the CARLA simulator road image well, [2]in the process of showing that the simulated driving result also approaches reality when the simulation image is realistic. [3] However, these FID scores are based on pretraining the inception v3 model with ImageNet. [4] Considering the autonomous driving algorithm, where segmentation information such as whether the pixel in the image is a road, lane, or car is important, the FID score based on the classification model can be considered inappropriate. Therefore, we propose a new method for measuring the Sim2Real gap in autonomous driving simulation images based on the lane segmentation model. Just as FID measures the difference in the distribution of the output vectors of the Inception model, it measures the difference in the attention maps from each encoder block in ENet-SAD. In this way, based on lane and road background information, etc., the similarity with the training dataset domain of ENet-SAD is quantitatively evaluated. At this time, by measuring the similarity of the average histogram of the attention map, the score between image bundles or videos can be measured as in FID. By applying this metric to our 'CARLA imageset with DCLGAN (hereafter referred to as CwD)' and 'CARLA imageset without DCLGAN (Cw/oD)', for which relative realism has already been quantified, it was confirmed that the similarity between the simulation images and the real images could be measured well.

## 2 Related Works

### 2.1 FID

The FID score uses the activation result vector of the last global spatial pooling layer after removing the output layer of the pre-trained Inception v3 model. Each image is represented by 2048 activation features, and FID measures the statistical similarity between two Gaussian groups of these feature vectors.[4] For the equation (1) $\mu$ means variance of features and $\Sigma$ means covariance. Importing feature vectors from other intermediate layers may change the number of features. FID is used as an index to evaluate the quality and realism of the generated image by measuring the distribution difference between the image generated by GAN and the real image. [6]

$$FID(T,G) = \|\mu_T - \mu_G\|^2 + Tr(\Sigma_T + \Sigma_G - 2(\Sigma_T\Sigma_G)^{\frac{1}{2}} \quad (1)$$

## 2.2 Attention Map

An attention map is a visual representation of the areas within an input image that are most relevant to a given task. Attention maps are commonly used in convolutional neural networks (CNNs) to identify which parts of an input image the network is focusing on while making predictions. The attention map can be calculated by analyzing the weights assigned to each input feature by the CNN's attention mechanism, which typically involves a combination of convolutional and pooling layers. This information can be used to improve the accuracy and interpretability of CNNs, as well as to identify potential biases or limitations in their performance. In addition, attention maps can be used to generate visual explanations for the network's predictions, which can be helpful in applications where transparency and interpretability are important. [7]

## 2.3 ENet-SAD

ENet-SAD is a lightweight model of ENet, which performs semantic segmentation, using Self Attention Distillation (SAD). SAD first extracts a self-attention map representing which part of the input image is focused on in each layer and minimizes the difference between the attention map of the next layer (teacher model) and the attention map of the current layer (student model). That is, knowledge distillation is performed from the back layer to the front layer within the same model. [4]

The expression to generate the attention map in ENet-SAD is as follows.

$$\Psi(.) = \Phi\left(\beta(\mathcal{G}^2{}_{sum}(.))\right). \tag{2}$$

$$\Phi(y_i) = \frac{e^{y_i/T}}{\sum_j e^{y_i/T}} \tag{3}$$

$\Phi$ is spatial softmax, $\beta$ is bilinear upsampling to match the attention map size, and $\mathcal{G}^2{}_{sum}$ is an attention mapping function obtained by squaring the sum of activation outputs for each channel of the input image. T in $\Phi$ is temperature.

## 2.4 Sim2Real gap in CARLA Simulator

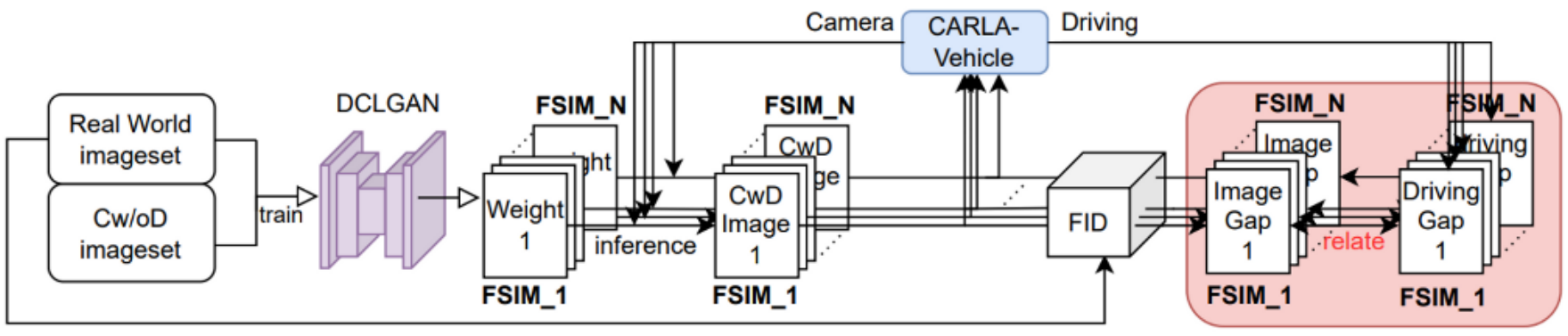

**Fig. 1.** Full framework for identifying Sim2Real image gaps and driving gaps from the CARLA-Vehicle-GAN loop

In our previous research [2], it was observed that the more realistically the image received from the camera attached to the vehicle of the CARLA Simulator became, the more realistically the driving result measured by GPS orbit or lane maintenance ability became. The CARLA image is converted to close to the real image by DCLGAN, and the realism of the image is measured by FID score. As a result, the FID value of CwD using DCLGAN was lower than that of Cw/oD without using DCLGAN. This paper quantitatively evaluated the Sim2Real gap of the input and output of the autonomous driving algorithm in the simulation and showed correlation, but it was not precisely checked whether the FID score was suitable for measuring the Sim2Real of the autonomous driving simulation image.

### 2.5 Histogram Comparison

Histogram is a graphical representation of the distribution of pixel intensities in an image which quantifies the number of pixels with each intensity value. Histograms are often used in image processing and analysis tasks such as image segmentation, where the goal is to partition an image into regions with similar pixel intensities.

To compare two histograms of the images, which are $H_1, H_2$ below equation (4), a metric $d(H_1, H_2)$ must be chose to express how much these histograms are similar with $N$ which is the total number of bins. [8]

$$d(H_1, H_2) = \frac{\sum_I (H_1(I) - \overline{H}_1)(H_2(I) - \overline{H}_2)}{\sqrt{\sum_I (H_1(I) - \overline{H}_1)^2 (H_2(I) - \overline{H}_2)^2}}, \text{ where } \overline{H}_k = \frac{1}{N} \sum_J H_k(J) \quad (4)$$

## 3 Approaches

FID score can measure how similar an image is to a real image based on the Inception v3 model pre-trained for classification of ImageNet images. However, it has low accuracy for images of unlearned classes and does not reflect spatial arrangement or semantic content well. Most of the images of the autonomous driving simulation are composed of roads, and most of these road images include lanes. The location, thickness, and sharpness of these lanes have a great influence on the Lane Keeping Assistant System (LKAS) algorithm. In fact, in a previous paper [3], it was confirmed that this information affects not only lane detection performance but also lane keeping ability during driving. Considering this, the semantic segmentation information of roads and lanes is important when measuring the reality of simulation road image sets for testing autonomous driving algorithms.

Therefore, in this paper, Sim2Real is measured by comparing the intermediate output of ENet-SAD, just as FID compares the output feature vector of the intermediate layer as shown in Fig. 2, but the attention map containing spatial information is used.

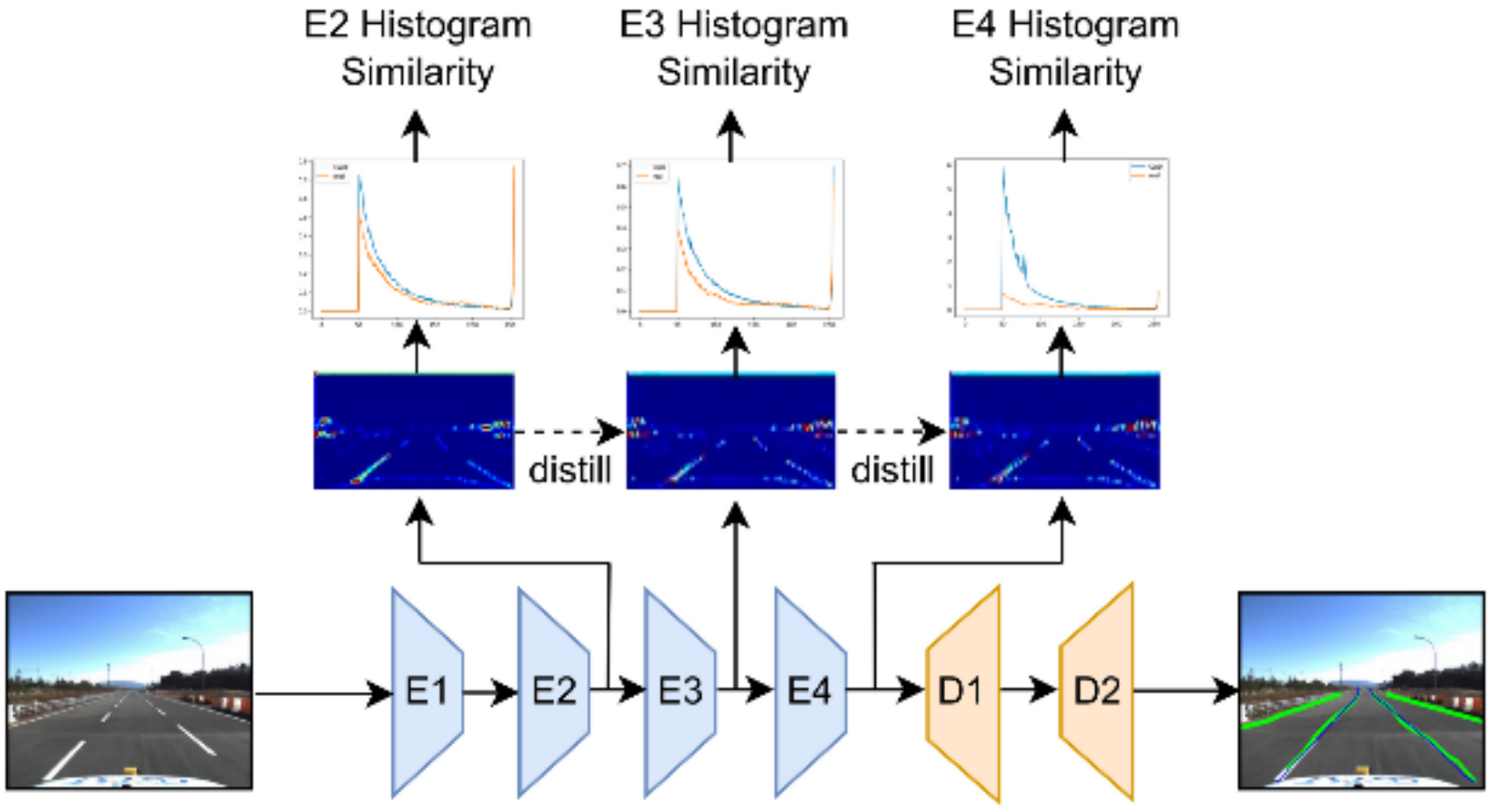

**Fig. 2.** overview of autonomous simulation image similarity measurement using ENet-SAD's attention map

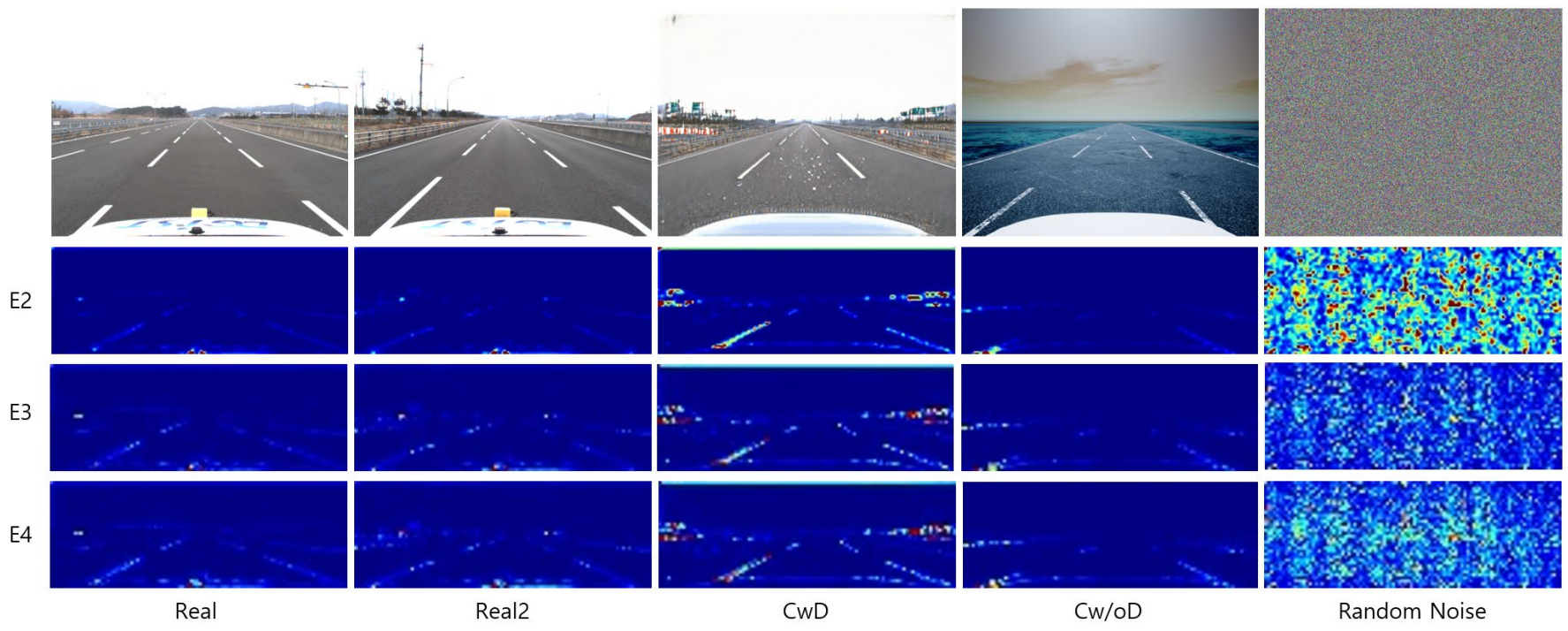

**Fig. 3.** Attention Map in each encoder from input images

The process of measuring similarity by using the attention map is shown in Fig. 2. The attention map is extracted through the attention generator function from three encoders (E2, E3, E4) in five cases as shown in Fig. 3. 'Real' is real world autonomous test road imageset in KIAPI(Korea Intelligent Automotive Parts Promotion Institute). ‘Real2’ is the other real world imageset to compare to ‘Real’. ‘CwD’ and ‘Cw/oD’ are imagesets in CARLA Simulator. We also added ‘Random Noise’ imageset to make a comparative dataset.

In the process of calculating the attention map using Equation (2), we set the temperature value of the spatial softmax function represented by Equation (3) to 10. This is done to flatten the intensity distribution of the attention map and prevent data loss when quantizing from 0 to 255 during the switch to a histogram. The exponential term in the softmax function can introduce significant deviation, which makes this precautionary measure crucial for accurate representation of the attention map.

After creating a histogram from this attention map, we excluded values from 0 to 100 on the x-axis for more efficient comparison considering that most of the intensity is concentrated at 0. If there are multiple images, the average histogram was obtained by accumulating each y value of the histogram and dividing it by the number of images. Afterwards, the similarity between each imageset was measured using Equation (4).

# 4 Results

The aim of this study was to evaluate the Sim2Real gap in autonomous driving road simulation images using attention map similarity, so we present the results of our study on quantifying the Sim2Real image gap in autonomous driving simulations using attention map similarity shown in Table.1. We have analyzed the similarity of attention maps across different datasets, which include Real to Real2, Real to CwD, Real to Cw/oD, and Real to Random noise. The similarity scores were computed using histogram comparison methods.

## 4.1 Real to Real2

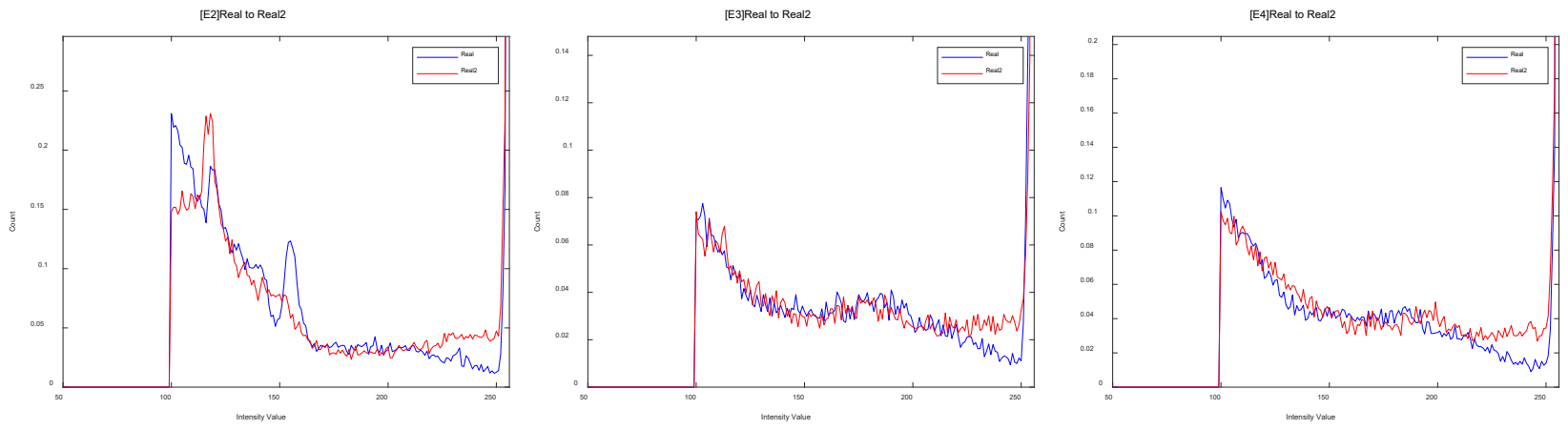

**Fig. 4.** Histogram similarity comparison graphs of Real to Real2

The similarity scores for Real to Real2 data indicate a high level of correspondence between the real-world data and the other real data. The scores were as follows:

0.9614 for the second-layer encoder, 0.9760 for the third-layer encoder, and 0.9887 for the fourth-layer encoder. These results suggest that the Real2 dataset is highly effective at replicating real-world scenarios.

### 4.2 Real to CwD

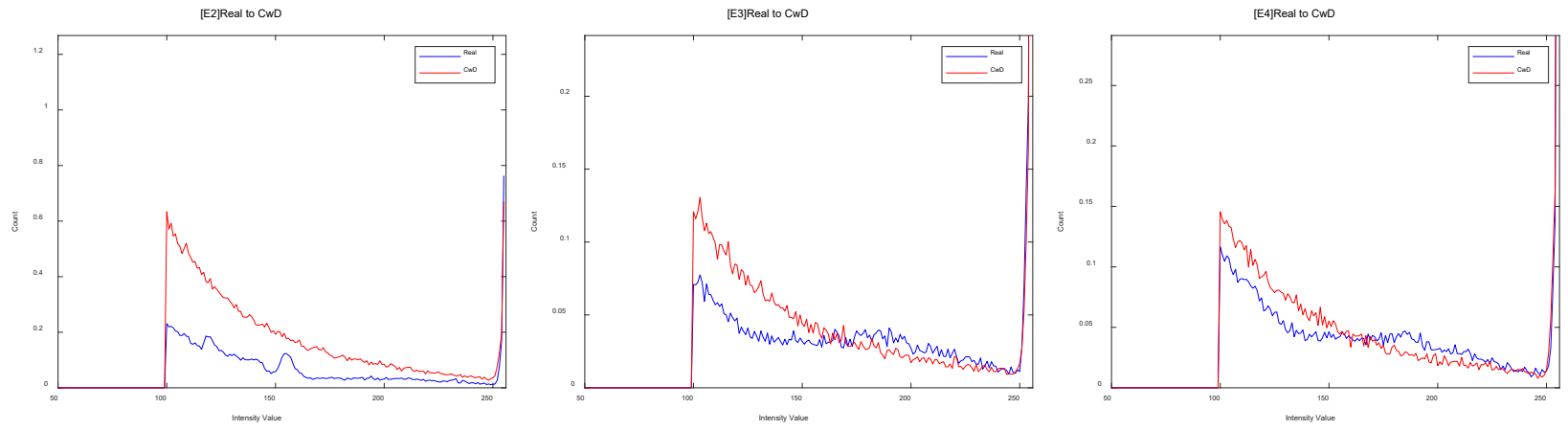

**Fig. 5.** Histogram similarity comparison graphs of Real to CwD

The similarity scores for Real to CwD data were 0.8899 for the second-layer encoder, 0.9585 for the third-layer encoder, and 0.9722 for the fourth-layer encoder. These results indicate that the DCLGAN used in the CwD is successful at bridging the Sim2Real image gap to a considerable extent.

### 4.3 Real to Cw/oD

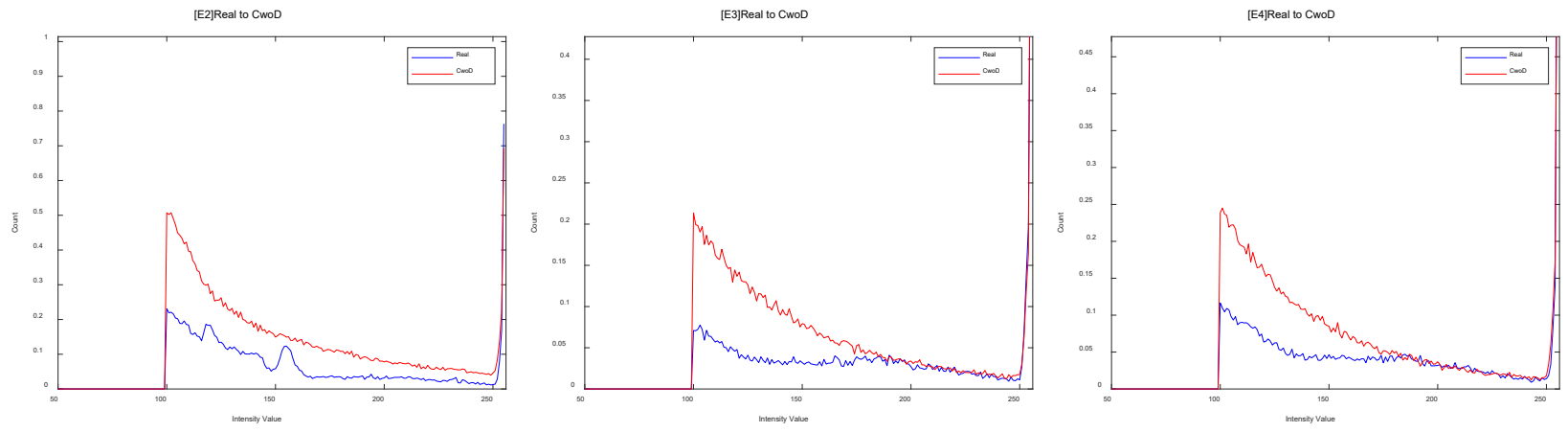

**Fig. 6.** Histogram similarity comparison graphs of Real to Cw/oD

The similarity scores for Real to Cw/oD data were 0.9157 for the second-layer encoder, 0.8641 for the third-layer encoder, and 0.8823 for the fourth-layer encoder. These results show that the Cw/oD dataset, which lacks domain adaptation, has a wider Sim2Real image gap compared to the CwD dataset.

### 4.4 Real to Random Data

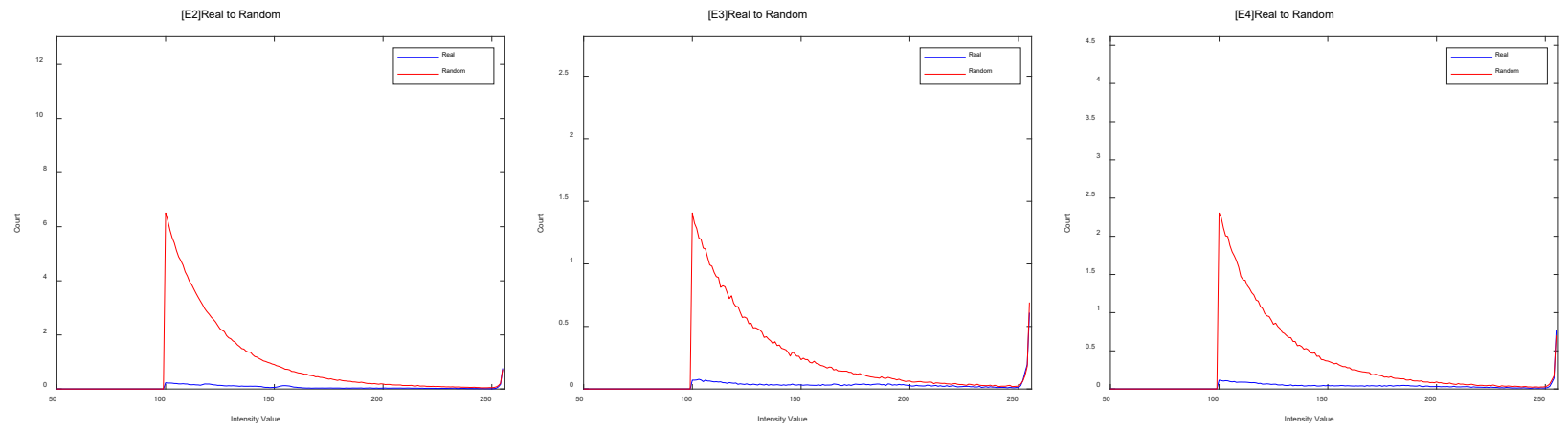

**Fig. 7.** Histogram similarity comparison graphs of Real to Random noise

The similarity scores for Real to random data were significantly lower than the scores for the other data sets, with 0.7264 for the second-layer encoder, 0.4508 for the third-layer encoder, and 0.4957 for the fourth-layer encoder.

### 4.5 Reliability of Attention Maps

The observed difference in similarity scores between CwD data and Cw/oD data in the second-layer encoder (0.8899 vs. 0.9157) can be attributed to the reliability of the attention map. Why attention map in the second-layer encoder is not as reliable as the ones in the third and fourth-layer encoders, is second-layer encoder is less trained for having attention on the lanes. That cause the results in a lower similarity score for the CwD data in this layer.

### 4.6 Similarity Analysis

Our results demonstrate the effectiveness of using attention map similarity to quantify the Sim2Real image gap in autonomous driving simulations. The high similarity scores for Real2 and CwD datasets indicate that these simulation techniques are successful at replicating real-world scenarios, while the lower scores for Cw/oD and random datasets emphasize the importance of domain adaptation in achieving realistic simulations.

**Table 1.** Results of Histogram Comparison by Encoders

| ImageSet | Similarity | | |
|---|---|---|---|
| | E2 | E3 | E4 |
| Real-Real2 | 0.9614 | 0.9760 | 0.9887 |
| Real-CwD | 0.8899 | 0.9585 | 0.9722 |
| Real-Cw/oD | 0.9157 | 0.8641 | 0.8823 |
| Real-Random | 0.7264 | 0.4508 | 0.4957 |

# 5 Conclusion

In this paper, we proposed a metric that quantifies the Sim2Real gap of images of autonomous driving simulations instead of FID. To this end, the histogram similarity of the middle self-attention map of ENet-SAD including spatial information was measured, and it was used to measure the similarity between real-real, real-realistically converted simulation images, real-vanilla simulation images and real-random images. We show that this measurement works well especially for the last encoder's attention map. This methodology has a limitation in that it has been tested only on road images without vehicles. In the future, we will experiment with situations where vehicles block lanes and seek ways to expand to models that include vehicle recognition information.

**Acknowledgments** This work was partially supported by the DGIST R&D Program of the Ministry of Science and ICT of Korea (23-IT-03)